\documentclass{article}
\usepackage{fullpage}
\usepackage{xcolor}         
\usepackage{microtype}
\usepackage{graphicx}
\usepackage{natbib}
\usepackage{amssymb,amsmath,amsthm}
\usepackage{subfigure}
\usepackage{booktabs} 

\newtheorem{theorem}{Theorem}
 
\DeclareMathOperator*{\argmax}{arg\, max} 
\usepackage{hyperref}
\usepackage[ruled,vlined,algo2e,linesnumbered]{algorithm2e}
\begin{document}

\title{Finite Sample Complexity Analysis of Binary Segmentation}
\author{Toby Dylan Hocking --- toby.hocking@nau.edu}
\maketitle

\begin{abstract}
Binary segmentation is the classic greedy algorithm which recursively splits a sequential data set by optimizing some loss or likelihood function.
Binary segmentation is widely used for changepoint detection in data sets measured over space or time, and as a sub-routine for decision tree learning.
In theory it should be extremely fast for $N$ data and $K$ splits, $O(N K)$ in the worst case, and $O(N \log K)$ in the best case. 
In this paper we describe new methods for analyzing the time and space complexity of binary segmentation for a given finite $N$, $K$, and minimum segment length parameter.
First, we describe algorithms that can be used to compute the best and worst case number of splits the algorithm must consider.
Second, we describe synthetic data that achieve the best and worst case and which can be used to test for correct implementation of the algorithm.
Finally, we provide an empirical analysis of real data which suggests that binary segmentation is often close to optimal speed in practice.
\end{abstract}

\section{Introduction}
\label{sec:introduction}

Data sequences are measured over time or space in many fields of study such as genomics \citep{olshen2004circular}, oceanography \citep{killick2010detection}, finance \citep{fryzlewicz2014wild}, and neuroscience \citep{Jewell2019}. 
In the analysis of such data, an important step involves detection of abrupt changes in the data distribution over time or space.
In this context, dynamic programming algorithms can be used to compute a globally optimal set of changes and segment-specific parameters for some loss or likelihood (even though the optimization problem is non-convex). 
For computing a model with $K$ splits for $N$ data, classic dynamic programming algorithms include the $O(K N^2)$ time algorithm of \citet{segment-neighborhood} and the $O(N^2)$ algorithm of \citet{Jackson2005}.
Since these classic algorithms are too slow for large $N$, new algorithms using pruning rules have been proposed to decrease computation time to $O(N\log N)$, while maintaining optimality \citep{Maidstone2016}. 
However even these new optimal algorithms are not fast enough for some applications, so an alternative is the classic binary segmentation heuristic \citep{binary-segmentation}.
Binary segmentation is not guaranteed to compute an optimal set of changepoints, but it can be extremely fast.
An entire regularization path of models from 0 to $K$ splits can be computed in worst case $O(N K)$, best case $O(N\log K)$ time.
Our paper provides a new method for analyzing the amount of computation that binary segmentation must do for a finite $N$ and $K$ (in the best and worst cases, as well as in real data sets).

\begin{figure}
    \centering
    \includegraphics[width=\textwidth]{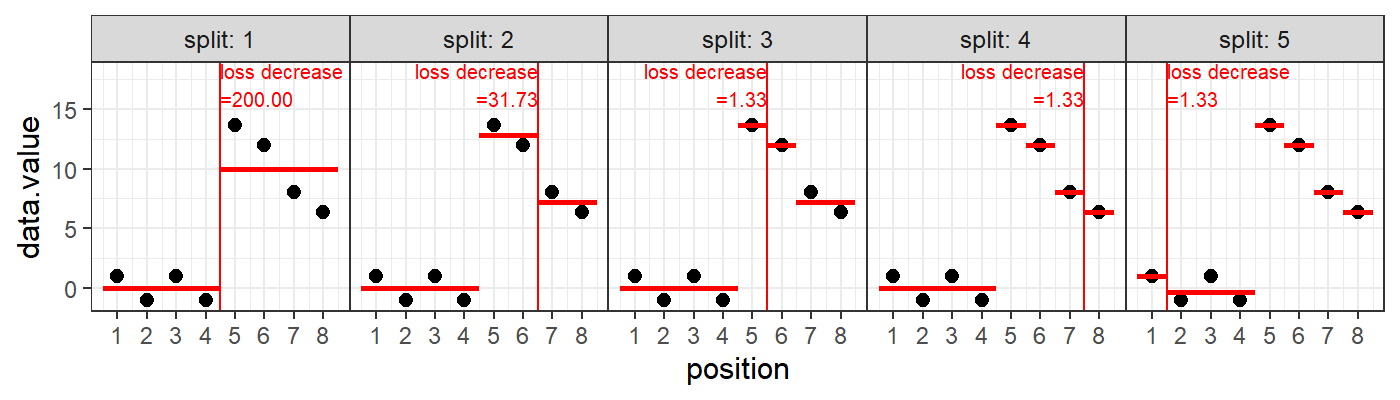}
    \vspace{-0.7cm}
    \caption{Demonstration of binary segmentation algorithm on a simple synthetic data set (details in Section~\ref{ties-in-different-segments}) for which a tie-breaking rule is required to achieve the best case number of splits to consider.
    After the first two splits, the next three splits all have an equal loss decrease value (1.33), so the best case time complexity can be achieved with a tie-breaking rule that chooses a split which results in the smallest number of candidate splits to consider afterwards.}
    \label{fig:synthetic}
\end{figure}

\subsection{Related work}
\label{sec:related-work}

The classic binary segmentation heuristic algorithm is originally attributed to \citet{binary-segmentation}.
It involves recusively searching a sequential data set for the the best split point, given previously chosen split points (Figure~\ref{fig:synthetic}).
More recent variants of binary segmentation have been proposed, such as searching a random subset of possible candidates in order to reduce computation time \citep{fryzlewicz2014wild, baranowski2019narrowest}.
Another idea motivated by reducing computation time is to limit the search to one changepoint each in a number of pre-defined seed intervals \citep{kovacs2020seeded}.
Because a known limitation is detection of short segments in longer ones \citep{venkatraman1992consistency}, more computationally intensive variants of binary segmentation have been proposed which involve searching for two split points simultaneously \citep{levin1985cusum,olshen2004circular}.
Binary segmentation is also an important sub-routine in decision tree learning algorithms such as CART \citep{breiman1984classification} and Maximum Margin Interval Trees \citep{Drouin2017}.
In each of the above algorithms there is a model complexity parameter (number of splits/segments), which needs to be appropriately chosen given the data.
This model selection problem can be solved using several unsupervised \citep{Yao88, mBIC} as well as supervised methods which can be used in situations where labeled training data are available \citep{Hocking2013icml, truong2017penalty}.

The dynamic programming algorithm we propose to compute the best case number of splits for binary segmentation is similar to the idea of the optimal binary search tree, for which there are many classic algorithms \citep{NAGARAJ19971}. 
For example \citet{Knuth1971} described $O(n^3)$ and $O(n^2)$ optimal algorithms for a tree with $n$ nodes, as well as $O(n\log n)$ heuristic/approximate algorithms which are very fast but not guaranteed to compute the optimal tree.
Later \citet{mehlhorn1975nearly} proved that the important heuristic was bisection, which always produces nearly optimal trees. 

\subsection{Contributions and organization}

Our main contributions are new algorithms for computing the best and worst case number of splits in binary segmentation, which allow finite sample time and space complexity analysis.
In Section~\ref{sec:bs-algo} we describe the classic binary segmentation algorithm, and provide new synthetic data examples that can be used for testing that the algorithm is correctly implemented.
In Section~\ref{sec:complexity-analysis} we propose new algorithms which can be used to compute the best and worst case number of splits for finite sample complexity analysis.
To the best of our knowledge this is the first time such finite sample complexity analysis has been described for binary segmentation with a minimum segment length parameter.
In Section~\ref{sec:results} we provide an empirical study of binary segmentation in several synthetic and real data sets.
Section~\ref{sec:discussion} concludes with a discussion of the significance and novelty of our findings.

\section{Binary Segmentation Algorithm}
\label{sec:bs-algo}

In this section we provide an overview of how the binary segmentation algorithm works. 
We assume there is a sequence of $N$ data for which we want to compute $K$ splits/iterations (max number of segments is $K+1$). 
We further assume that each iteration splits a segment into two segments, each of minimum size $m$ (this is the minimum segment length parameter).
Finally, we assume there is a loss function $\ell$ that binary segmentation seeks to minimize (for example, common choices are the square loss or L1 loss for real-valued data, and the Poisson loss for non-negative integer count data).
The binary segmentation algorithm performs the following computations: 
(a) for any new segments, find the split which results in the best loss decrease; (b) store that segment/split in a container for future lookup; 
(c) look in that container for the  segment with best loss decrease and split it into two new segments; 
(d) repeat until the max number of splits $K$ is reached.
To analyze the time complexity of step (a) we use the number of candidate splits that must be considered to find the best loss decrease (Table~\ref{tab:best-worst-time}).
Note that this is a valid choice for finite sample time complexity analysis since the loss of each candidate split can be computed in amortized constant time for common loss functions including square/L1/Poisson.
The positions after which a change may occur in $N$ data are $\{m,\dots,N-m\}$ and the size of this set (number of candidate splits that must be considered) is
\begin{equation}
    \label{eq:g}
    g(N) = (N-2m+1)_+.
\end{equation}
To analyze the time and space complexity of steps (b) and (c) we need to specify what type of container is used. 
For efficiency the container should be a red-black tree or similar data structure which allows for constant time lookup of the item with best loss decrease, and insertion of a new item which is logarithmic in the container size.
For finite sample complexity analysis of steps (b) and (c) we therefore need to examine the container size before each insert (Table~\ref{tab:best-worst-storage}).
Note that in both steps (a) and (b) the ``best'' loss decrease must be identified.
In order for the algorithm to achieve the best case number of candidate splits considered, it must correctly perform tie-breaking, as explained in the next section.

\begin{table}[t]
    \centering
    {\small
\begin{tabular}{cc|cc|cc}
&& \multicolumn{2}{c|}{Best case: equal splits} &
\multicolumn{2}{c}{Worst case: unequal splits} \\
\hline
$j$ & $I$ & this iteration & total & this iteration & total \\
\hline
1 & 1 & $N-1=63$  & $N-1=63$    & $N-1=63$ & $N-1=63$ \\
2 & 2 & $N-2=62$  & $2N-3=125$  & $N-2=62$ & $2N-3=125$ \\
  & 3 & $N/2-2=30$&             & $N-3=61$ & $3N-6=186$ \\
3 & 4 & $N/2-2=30$& $3N-7=185$  & $N-4=60$ & $4N-10=246$ \\
$\vdots$ &     $\vdots$ &     $\vdots$  & $\vdots$ & $\vdots$ & $\vdots$\\
4 & 8 & $N/4-2=14$& $4N-15=241$ & $N-8=56$ & $8N-36=476$ \\
$\vdots$ &     $\vdots$ &     $\vdots$  & $\vdots$ & $\vdots$ & $\vdots$ \\
7 & 64 & $N/32-2=0$&$7N-128=321$ & $N-64=0$ & $64N-2080=2016$ \\
\end{tabular}
}
    \caption{Counts of best and worst case number of candidates to compute for $N=64$ data and min segment length $m=1$.
    Each row shows the number of splits for a given iteration $I$ and tree depth $j$, along with the cumulative total.}
    \label{tab:best-worst-time}
\end{table}

\subsection{Tie-breaking rules and test cases}

In this section we provide examples of simple synthetic data which have splits of equal loss, and thus require a tie-breaking rule in order to achieve the best case number of splits.
We begin by discussing the case of ties in split points within a segment, then we discuss the more complex case of ties in different segments that could be split.

\paragraph{Ties in split points on a segment.} 
When the binary segmentation algorithm considers a new segment to split, it must optimize over all possible split points, by taking the split that results in minimum loss. 
If there are several splits which each have the same minimum loss value, then to minimize the number of candidates to optimize over in the next step, the binary segmentation algorithm should first break ties using the number of split candidates which would need to be computed, then the distance from the start/end of the segment. 
A simple example is the data sequence [1,2,...,8] with the L1 loss and min segment length $m=1$.
The optimal L1 loss values for the seven possible splits are [12,10,8,8,8,10,12], so there are three splits which result in the same min loss.
Each of these three splits has the same number of split candidates which would need to be computed during the next iteration, $g(3)+g(5)=g(4)+g(4)=6$.
The best case number of splits for future iterations is achieved if the algorithm chooses the split in the middle (two child segments of size 4).
More generally, let $\underline p$ be the segment start, $\overline p$ be the segment end, and $c\in\{\underline p, \dots, \overline p-1\}$ be a candidate split point.
The distance from split $c$ to the nearest start or end is
\begin{equation}
    D_c = \min\{ c-\underline p, \overline p-c-1 \}.
\end{equation}
Let $\mathcal C$ be the set of splits to tie-break, with min loss and min number of split candidates which would need to be computed during the next iteration.
The best case tie-breaking rule chooses the split with maximal value for the distance to nearest start or end,
\begin{equation}
    c^* = \argmax_{c\in\mathcal C} D_c.
\end{equation}
In the example above, the segment start $\underline p=1$, end $\overline p=8$, and set of splits to tie-break $\mathcal C=\{3,4,5\}$. 
Distances are $D_3=2$, $D_4=3$, $D_5=2$, so the best split is therefore $c^* = 4$, which results in two segments of size 4.
This requires computing six new candidates in the next step (three on each of the two segments with four data), then four in the following steps (one on each of the four segments with two data), for a total of 10 candidates.
This is preferable to the other options, which result in a larger number of candidates (4 and 2 with segments of size 3 and 5; then 1,2,2 for segments of size 2,2,3; then 1 for final segment of size 2; total number of candidates 12).

\begin{table}[t]
    \centering
    {\small
\begin{tabular}{c|cc|cc}
     & \multicolumn{2}{c|}{Equal splits}    & \multicolumn{2}{c}{Unequal splits}  \\
     \hline
     &  sizes before  &   size after   & sizes before & size after \\
 iteration $I$ &  inserts &  iteration &  inserts &  iteration \\
\hline
 1 & 0   & 1 & 0 & 1 \\
 2 & 0,1 & 2 & 0 & 1 \\ 
 3 & 1,2 & 3 & 0 & 1 \\
$\vdots$ &     $\vdots$ &     $\vdots$  & $\vdots$ & $\vdots$ \\
32 & 30,31 & 32 & 0 & 1 \\
33 &       & 31 & 0 & 1 \\
34 &       & 30 & 0 & 1 \\
$\vdots$ &     $\vdots$ &     $\vdots$  & $\vdots$ & $\vdots$ \\
62 &       & 2 & 0 & 1 \\
63 &       & 1 & 0 & 1
\end{tabular}
}
    \caption{Table of best and worst case storage, for $n=64$ data with min segment length $m=1$ and max number of segments/splits.
    Sizes before inserts columns show the size of the container before each insert operation (to analyze time complexity of the insert operation); size after iteration columns show the size of the container after each iteration (to analyze space complexity).
    Equal splits results in largest container size before inserts of $O(n)$, and worst case space complexity of $O(n)$. 
    Unequal splits results in smallest container size before inserts (always zero), and best case storage (container size 1 after each iteration).}
    \label{tab:best-worst-storage}
\end{table}

\paragraph{Ties in different segments that could be split.} \label{ties-in-different-segments}
A flexible and efficient implementation of binary segmentation requires a container to store the different segments that could be split during each iteration. 
The segments in this container should be sorted by loss decrease values; the next segment to split should result in the largest loss decrease.
Similar to the discussion above about ties in split points on a segment, is is possible for there to be several segments which have the same loss decrease value. 
In this case, to achieve the best case number of splits to optimize over in subsequent iterations, the algorithm should again break ties in the same way. 
A simple synthetic data set for which this tie-breaking rule is necessary using the L2 square loss can be constructed in the following manner. 
First, let $a,b,x,\epsilon\in\mathbb R$ be real number parameters which will determine our synthetic data  $[a,-a,a,-a|b+\epsilon+x,b+\epsilon|b-\epsilon,b-\epsilon-x]$.
The idea is to choose $b$ and $\epsilon$ large enough so that the first two splits happen at the vertical bar positions, resulting in three segments.
Then $a$ and $x$ can be chosen such that all three segments have an equal value for the best loss decrease. 
We chose $a=1$, $b=10$, $x=\sqrt{8/3}$, $\epsilon=2$, which results in the data shown in Figure~\ref{fig:synthetic}.
The algorithm proceeds as follows:
\begin{description}
\item[Split 1:] 7 candidates are considered, best has a change after position 4. 
\item[Split 2:] Each of the two new children segments created from split 1 require computing loss of 3 split candidates, for a total of 6 candidates considered.
Splitting optimal segment results in a change after position 6, whereas the other segment with loss decrease of 4/3 stays in the container. 
\item[Split 3:] Each of the two new children segments created from split 2 require computing loss of 1 split candidate, for a total of 2 candidates considered.
Both new segments have loss decrease of 4/3, so container now has three segments with the same loss decrease value.
Tie-breaking is performed by minimizing the number of split candidates which would be required to compute in the next iteration; the result is a change after position 5.
Two segments with the same loss decrease value remain in the container.
\item[Split 4:] Both of the two new children segments created from split 3 are of size 1, so can not be split, and do not require any new split candidates to be computed.
Next split results in a change after position 7, and one segment remains in the container.
\item[Split 5:] Again no new split candidates are computed, and instead the last segment in the container is split, resulting in a change after position 1, and an empty container.
\end{description}
Looking at the example above, we see that the total number of split candidates considered is 15, which is indeed the best case for $N=8$ data, $K=5$ splits, and a min segment length of $m=1$.
In contrast if we had not implemented tie-breaking, and had instead chosen split 3 after position 1, then we would have computed two additional split candidates, for a total of 17.
The worst case would be achieved for the data sequence [-1,1,-1,1,-1,1,-1,1] which for $K=5$ splits would require computing the loss for a total of 25 split candidates. 
In the next section we describe methods that can be used to compute the best/worst case number of iterations for any data size $N$, number of splits $K$, and min segment length $m$.

\section{Finite Sample Complexity Analysis}
\label{sec:complexity-analysis}

We begin this section with an analysis of the number of candidate splits that must be considered when computing the maximum number of splits, $K=N-1$. 
For simplicity we first assume a min segment length of $m=1$ and that $N$ is a multiple of 2; we can let $N=2^J$ for some $J\in\{1,2,\dots\}$, for example $J=6
  \Rightarrow N=64$.
Then for any $j\in\{1,\dots,J+1\}$ if we do $I=2^{j-1}$ iterations/splits then the number of candidates that must be considered is shown in Table~\ref{tab:best-worst-time}.
In the best case, each split results in segments of equal size, and we have $Nj -2^j + 1 = N(1+\log_2 I) -I/2 +1 \Rightarrow O(N \log I)$ total candidates to consider.
In the worst case, each split results in segments of unequal sizes, and we have $NI - I(1+I)/2 \Rightarrow O(NI)$ total candidates to consider.
 
\begin{figure}[t]
    \centering
    \includegraphics[width=\textwidth]{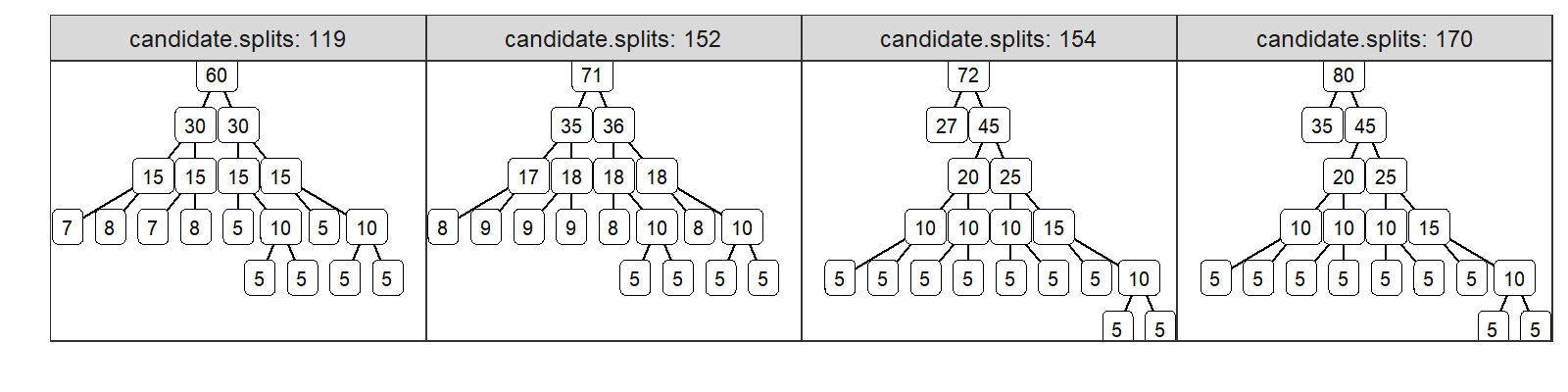}
    \vspace{-1cm} 
    \caption{Optimal binary trees constructed to determine optimal number of candidate splits for $N\in\{60,71,72,80\}$ data, min segment length $m=5$, and number of splits/iterations $I=9$.
    Smaller $N$ values result in a balanced first split, whereas larger $N$ values result in an unbalanced first split (one small child with no splits, one large child with all remaining splits).}
    \label{fig:optimal-trees-some}
\end{figure}
To store previously computed best loss/split for each segment, we use a container with insert operations which are logarithmic in container size ---
if the container has $p$ items then an insert takes $O(\log p)$ time. 
In Table~\ref{tab:best-worst-storage} we therefore analyze the container size in the two extreme cases (equal and unequal splits). 
The ``sizes before inserts'' columns show $p$ before each insert (to analyze time complexity), and the ``size after iteration'' columns show $p$ after inserts (to analyze space complexity). 
For unequal splits, the total time is linear, and the total space is constant, because the container always stays at size 1.
For equal splits, the total time over all inserts is $-\log(I-1) + \sum_{p=1}^{I-1} 2\log p\in O(I\log I)$, which is smaller than the $O(N\log I)$ time for the split computation, so the container insert step is not a limiting factor for time complexity.
However for equal splits, the space complexity achieves the worst case, a max container size of $N/2$ segments for which we have already computed the best split point, but we have not yet split, at iteration $N/2$.

\subsection{New dynamic programming algorithm for computing best case number of candidate splits}
\label{sec:time-complexity}

In the previous section we provided an overview of how the finite sample complexity analysis works for the special case of the data size being a multiple of 2.
The worst case analysis (counting number of candidates for unequal splits) also works for the general case, but the best case analysis does not (equal splits do not always result in the best case for small $K$).  
In this section we describe a new algorithm which can be used to compute the best case  number of splits that must be considered for any data size $N$, number of splits $K$, and min segment length $m$.

Recall from (\ref{eq:g}) that $g(N)$ is the number of candidate splits which must be considered in order to find the optimal split on the segment of size $N$.
Let $f(N,K)$ be the best number of candidates that must be computed if the segment of size $N$ is split $K$ times.
The geometric interpretation of this quantity involves a binary tree in which each node represents a segment with a certain size, and the size of each parent always equals the sum of sizes over the two children (Figure~\ref{fig:optimal-trees-some}). 
Computing $f(N,K)$ involves searching the space of all trees with root size $N$ and $K$ splits, in order to find the tree with min cost, as quantified by summing $g(s)$ over all node sizes $s$.

The key idea of our proposed algorithm is that $f(N,K)$ can be recursively computed.
For some ideal smaller number of changes $d\in\{0,\dots,K-1\}$, and for some ideal split size $s$, we have  $f(N,K) = g(N) + f(s,d) + f(N-s,K-d-1)$.
To compute this we need to search over all possible numbers of changes $d$ and split sizes $s$.
For any number of changes $d$, we have the following lower bound on the size $s$ due to the min segment length $m$,
\begin{equation}
    s \geq (d+1)m.
\end{equation}
For the same reason we have the following upper bound,
\begin{equation}
    s \leq N+(d-K)m.
\end{equation}
For example consider computing $f(15,2)$ with min segment length of $m=3$. 
In that case we need to search over $f(s,0)+f(15-s,1)$ for seven sizes $s\in\{3,4,5,6,7,8,9\}$.
Additionally when the number of changes in the two child segments is the same, the symmetry of the problem implies the following stronger upper bound,
\begin{equation}
    d = (K-1)/2 \Rightarrow s\leq \lfloor N/2 \rfloor.
\end{equation}
For example consider computing $f(15,3)$ with min segment length of $m=3$.
In that case we need to search over $f(s,0)+f(15-s,2)$ for four sizes $s\in\{3,4,5,6\}$, and $f(s,1)+f(15-s,1)$ for just two sizes $s\in\{6,7\}$.
When there are $d$ changes the overall upper bound on $s$ is therefore
\begin{equation}
    \overline s(N,K,d,m) = \begin{cases}
    N+(d-K)m & \text{ if } d < (K-1)/2,\\
    \lfloor N/2 \rfloor & \text{ if } d = (K-1)/2.
    \end{cases}
\end{equation}
For $N$ data, $K$ changes, min segment length $m$, and any smaller number of changes $d<K$, we therefore have the following set of sizes $s$ to optimize over,
\begin{equation}
    \mathcal S(N,K,d,m) = \{(d+1)m, \dots, \overline s(N,K,d,m)\}.
\end{equation}
The results above are combined into the theorem below, which gives an efficient and optimal method for computing the best case number of splits that need to be computed during binary segmentation.
\begin{theorem}
The best case number of candidate splits that must be computed in binary segmentation, if a segment of $N$ data is split $K$ times into segments of min size $m$, can be determined as follows.
First, for all $N$ we initialize $f(N,0) = g(N)$. 
Then, for all $K>0$ and for all $N$ we use the following dynamic programming update rule.
\end{theorem}
\begin{equation}
    \label{eq:f}
    f(N,K) = g(N) + 
    \min_{
    d\in\{0,\dots,\lfloor (K-1)/2\rfloor\}
    }
    \min_{
    s\in\mathcal S(N,K,d,m)
    }
    f(s,d) + f(N-s,K-d-1).
\end{equation}
\begin{proof}
The proof is straightforward, by induction on $N$ and $K$. 
\end{proof}

\begin{algorithm2e}[t]
\SetAlgoLined
 \caption{Computing the best case number of splits in binary segmentation}\label{algo:DP}
Input: Number of data $N$, number of changes $K$, min segment length $m$\\
Initialize data structure to store optimal cost values: $F$\\
\For{depth d from 0 to $K$}{ \label{line:for} 
  \For{size $s\in\mathcal S(N,K,d,m)$}{
    $\text{min\_cost}\gets\text{ if }d=0\text{ then } 0 \text{ else } \text{getMin}(s,d,m,F)$\\
    $F[s,d]\gets \text{min\_cost}+(N-2m+1)_+$
  }
}
Output: $F[N,c]$, smallest number of splits that must be considered.
\end{algorithm2e}
\begin{algorithm2e}[t]
\SetAlgoLined
 \caption{getMin sub-routine}\label{algo:getMin}
Input: Number of data $N$, number of changes $K$, min segment length $m$, optimal cost values $F$\\
Initialize $\text{min\_cost}\gets \infty$\\
\For{depth d from 0 to $\lfloor (c-1)/2\rfloor$}{
  \For{size $s\in\mathcal S(N,K,d,m)$}{
    $\text{cost}\gets F[s,d] + F[N-s, K-d-1]$\\
    if $\text{cost} < \text{min\_cost}$ then $\text{min\_cost}\gets\text{cost}$
  }
}
Output: min\_cost, smallest number of splits that must be considered.
\end{algorithm2e}

\paragraph{Pseudo-code and complexity analysis of proposed algorithm.} 
The dynamic programming algorithm is summarized in Algorithm~\ref{algo:DP}, which has two for loops (over depth and size). 
Inside each iteration of those two for loops is a call to the getMin sub-routine (Algorithm~\ref{algo:getMin}), which also has two for loops (over depth and size).
Overall the proposed dynamic progamming algorithm has time complexity of $O(N^2 K^2)$, so it only reasonable to run for relatively small number of data $N$ and splits $K$.
\section{Empirical results}
\label{sec:results}

In this section we analyze the results of running our proposed Algorithm~\ref{algo:DP}, and show how it can be used to compare the best case with the empirical number of candidate splits computed in real data sets.
All experimental results not depend on the particular computer hardware used; we used a laptop with a 2.40GHz Intel(R) Core(TM)2 Duo CPU P8600.

\subsection{Analysis of optimal binary trees}
In order to visualize the structure of the optimal binary trees when the number of splits $K$ is less than the number of data $N$, we ran Algorithm~\ref{algo:DP} using $N\in\{60,71,72,80\}$ data, min segment length $m=5$, and number of splits/iterations $I=9$.
We expected results qualitatively similar to the best case binary trees produced when the number of splits is set to the max value, $K=N-1$ (in these trees each split creates two children of equal size, as explained in Section~\ref{sec:complexity-analysis}).
Interestingly, we observed that this is only the case when $N$ is relatively small (Figure~\ref{fig:optimal-trees-some}, $N=60$ and 71). 
When $N$ is relatively large, we observed qualitatively different best case trees in which the first split creates one small node with no children, and one large node with the rest of the children (Figure~\ref{fig:optimal-trees-some}, $N=72$ and 80).
These results show that computing the tree in which children are always of nearly equal size is not sufficient (and is heuristic) for computing the best case number of candidate splits.

\subsection{Analysis of number of candidate splits considered in real data sets}

In this section we examine the number of candidate splits which must be computed in real data sets, in order to determine if binary segmentation is closer to the best or worst case in practical scenarios. 
We expected that binary segmentation should be closer to the best case in real data sets, since the worst case only happens with pathological data sets (such as data which bounce up and down).
Since the proposed Algorithm~\ref{algo:DP} is quadratic time, we only ran it for small data sizes, so we only know the true lower bound on the number of candidate splits for small data size.
For larger data sizes, we computed an inexact lower bound using a fast heuristic algorithm which performs a depth-first search of the binary tree with equal sized children.

\begin{figure}[t]
    \centering
    \includegraphics[width=\textwidth]{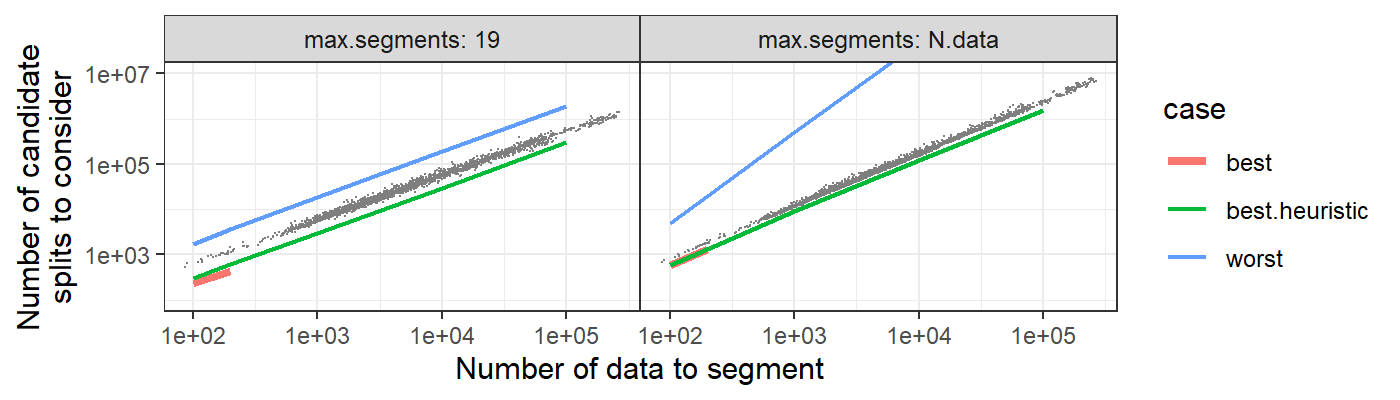}
    \vspace{-0.7cm}
    \caption{Analysis of 2752 real genomic count data sets from McGill benchmark of data size $N$ from 87 to 263169, using binary segmentation with the Poisson loss. 
    Number of candidate splits to consider in real data (grey dots) achieves the asymptotic best case, $O(N\log N)$.
    }
    \label{fig:mcgill-iterations}
\end{figure}

\paragraph{Peak detection benchmark.} 
We used data from the McGill ChIP-seq benchmark \citep{Hocking2017bioinfo}, which does not have a license but is freely downloadable from \url{https://rcdata.nau.edu/genomic-ml/chip-seq-chunk-db/}.
This benchmark consists of 2752 real genomic data sets of size $N$ from 87 to 263169.
Since these data are non-negative counts, we used the Poisson loss.
The main goal in analyzing these data is to determine the positions of peaks, which are regions where the signal jumps abruptly up and then back down again.
We therefore suspected that there may be some evidence of worst case number of candidate splits to consider (because the data bounce up and down). 
However, we observed that the empirical number of splits that the algorithm considered closely matches the best case (Figure~\ref{fig:mcgill-iterations}).
We used Algorithm~\ref{algo:DP} to compute best case number of candidate splits for $N=100$ to 200, and observed that the heuristic is exact when max segments equals number of data, but the heuristic is an over-estimate of the lower bound when max segments is smaller than the number of data (as expected, see discussion in Section~\ref{sec:complexity-analysis}).

\paragraph{Copy number benchmark.} 
We used data from R package neuroblastoma \citep{neuroblastoma}, which is freely downloadable from the Comprehensive R Archive Network (CRAN) under the GPL-3 license.
This benchmark consists of 13721 real genomic data sets of size $N$ from 11 to 5937, and we used the square loss since the data were real numbers.
In these data we used Algorithm~\ref{algo:DP} to compute exact lower bounds for all data sizes $N$ from 11 to 100. 
We observed that the heuristic was always exact when max segments equals number of data, and it was exact for five sizes, $N\in\{11,12,13,18,19\}$, when max segments was set to 10.
As in the previous benchmark, we observed that the empirical number of splits that the algorithm considered closely matches the best case (Figure~\ref{fig:mcgill-iterations}).
In fact we observed 45 real data sets for which the empirical number of splits considered was less than the heuristic/inexact lower bound.
Overall these data provide convincing evidence that the number of splits that binary segmentation considers in real data closely matches the theoretical best case.

\begin{figure}[t]
    \centering
    \includegraphics[width=\textwidth]{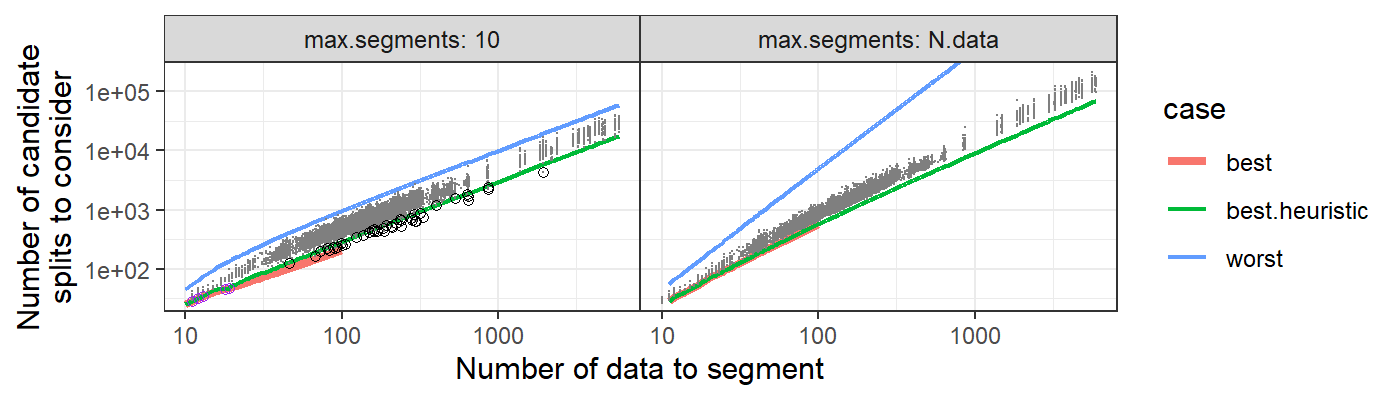}
    \vspace{-0.7cm}
    \caption{Analysis of 13721 real genomic data sets from neuroblastoma benchmark of data size $N$ from 11 to 5937, using binary segmentation with the square loss. 
    Number of candidate splits to consider in real data achieves the asymptotic best case, $O(N\log N)$, and in 45 instances (black circles) requires fewer candidate splits than predicted by the best case heuristic.
    For max segments = 10, we used dynamic programming to compute the best case number of splits for all data sizes between 11 and 100 (orange line), and the heuristic (green line) was exact for only 5 data sizes $N\in\{11,12,13,18,19\}$ (purple circles).  }
    \label{fig:neuroblastoma-iterations}
\end{figure}


\section{Discussion and conclusions}
\label{sec:discussion}
This paper presented new methods for analyzing the time complexity of the classic binary segmentation algorithm for changepoint detection.
Our work analyzed the speed of the classic algorithm in terms of the number of candidate splits that must be considered in each iteration of the algorithm, and in terms of the number of previously computed splittable segments that must be stored. 
We proposed a new dynamic programming algorithm for computing a lower bound on the number of candidate splits that must be considered, thereby providing a method for finite sample analysis of the time complexity of the algorithm. 
We analyzed several real data sets in terms of the number of candidate splits that were computed, and compared those empirical observations to the upper and lower bounds.
We observed that in those real data sets, the empirical number of candidate splits is only a constant factor larger than the lower bound, indicating that the binary segmentation algorithm achieves the asymptotic best case in practical real data scenarios.
Our analysis provides convincing evidence that binary segmentation can be quickly computed for large real data sets, and large model sizes.
Our work therefore suggests that the speed of binary segmentation is optimal in real data, and so recently proposed simplifications \citep{fryzlewicz2014wild, kovacs2020seeded} of the classic algorithm only result in constant factor speedups.
One drawback/limitation of our work was the time complexity of our proposed dynamic programming algorithm, which was quadratic in both number of data and splits.
For future work, we plan to investigate faster algorithms for computing the best case lower bound, using ideas from the optimal binary search tree literature \citep{NAGARAJ19971}.

\paragraph{Reproducible Research Statement.} 
An R package with C/C++ code that implements our proposed algorithm is available at \url{https://github.com/tdhock/binsegRcpp}.
In addition, the GitHub repository \url{https://github.com/tdhock/binseg-model-selection} contains code for reproducing the figures in this paper.

\bibliography{refs}
\bibliographystyle{abbrvnat}

\end{document}